\documentclass[twocolumn]{article}

\usepackage{arxiv}
\usepackage{xurl}
\usepackage[colorlinks=true,allcolors=blue]{hyperref}
\usepackage[utf8]{inputenc} 
\usepackage[T1]{fontenc}    
\usepackage{url}    
\usepackage{tabularx}
\usepackage{array}
\usepackage{booktabs}       
\usepackage{amsfonts}       
\usepackage{nicefrac}       
\usepackage{microtype}      
\usepackage{doi}
\usepackage[T1]{fontenc}
\usepackage{makecell}
\usepackage{amsmath}
\usepackage[numbers]{natbib}
\usepackage[capitalise,noabbrev]{cleveref}
\usepackage{graphicx}
\usepackage{cuted}
\usepackage{tabularx}
\usepackage{booktabs}

\graphicspath{ {./images/} }
\usepackage{xcolor}
\usepackage[acronym]{glossaries}
\usepackage{fancyhdr} 
\glsdisablehyper  
\usepackage{gensymb}
\usepackage{ulem}
\usepackage{subcaption}
\newcommand{\affit}[1]{$^{\mathrm{\textnormal{\textit{#1}}}}$}

\title{Post-Training Quantization of Autoregressive Weather Models}

\author{
 Ananyo Bhattacharya\affit{a} \\
\And
 Swastik Bhattacharya\affit{b} \\
\And
 Christiane Jablonowski\affit{a} \\
}

\begin{document}







\twocolumn[
\begin{@twocolumnfalse}
\maketitle

\begin{abstract}
  Advancements in high-resolution numerical weather prediction (NWP) and data assimilation (DA) have shaped the developments in deep learning (DL) architectures emulating atmospheric dynamics. Emulators for weather forecasting exhibit forecast quality comparable to physics based models at forecast horizon scaling from few days to subseasonal time scales. The emulators are driven by hardware-accelerated matrix multiplication in autoregressive inferences, significantly reducing the computation time and resources required for NWP. Optimization of the matrix multiplication processes in GPU architectures provides opportunities to scale towards high-resolution domain, and offers implementation of out of the box solutions. Post-training quantization (PTQ) has been demonstrated across multiple DL architectures to accelerate and increase the number of computations in unit time while consuming less power, enabling applications on edge hardware. In this study, we investigate the effect of PTQ on pre-trained AI emulators for global-scale weather forecasting. We implement PTQ algorithms in Deep Learning Weather Prediction (DLWP) and FourCastNet (FCN) models as a proof of concept for geophysical fluid dynamics applications. We systematically investigate the effect of PTQ on emulator inferences over short-range forecast horizons. Evaluation of PTQ configurations using simulated quantization hints at qualitatively meaningful forecasts over short-time horizons. These results provide a first benchmark of PTQ for autoregressive weather emulators and a basis for quantization-based optimization of DL models for dynamical systems.
\end{abstract}

\bigskip
\vspace{0.5cm}
\end{@twocolumnfalse}
]
\footnotetext[1]{University of Michigan, Ann Arbor, USA}
\footnotetext[2]{University of Texas at Dallas, Dallas, USA}

\section{Introduction}\label{sec1}

The estimation of atmospheric state variables requires a combination of numerical weather prediction and insights from real-world observations involving ground stations and satellite-derived data products. Deterministic weather forecasting requires initial conditions of the surface and atmospheric state variables to autoregressively estimate the future states. Even under consistent frameworks for solving the prognostic equations, the model solutions are highly sensitive to the prescribed conditions and assumptions. Therefore, the quality of the forecast degrades with time owing to chaotic interactions within the system \citep{lorenz2017deterministic, price2025probabilistic}. Combining real-world observations with physics-based models through data assimilation provides precise information on global and regional scale meteorology. Therefore, the quality of weather forecasting models inevitably influences the estimation of atmospheric state, cloud formation, and precipitation. \\

The time and computational costs associated with running NWP models pose a challenge to scaling at higher spatio-temporal resolution and their accessibility. Depending on the type of application, the physics-based Earth system models have been adapted to different spatial resolutions. The long-term climate studies like the Coupled Model Intercomparison Project (CMIP) generally run at a coarse spatial resolution of 100-200 km \cite{balaji2018requirements}, while the National Weather Service operates the High Resolution Rapid Refresh (HRRR) model \cite{dowell2022high} at 3 km resolution over the contiguous United States (CONUS) domain. Recent efforts are attempting to push the spatial scaling to a 1 km resolution global scale forecasting.\\

The availability of two and three-dimensional surface and atmospheric datasets at global and regional scales has enabled the application of deep learning (DL) architectures to weather and climate forecasting. Global reanalyses, such as ECMWF ERA5, together with historical climate datasets, provide long-term records of atmospheric processes that support the training of DL models to represent atmospheric dynamics at high spatio-temporal resolution. Using DL method reduces the computational costs with respect to using traditional NWP solvers. Recent advances in DL architectures and their application to geophysical fluid dynamics have enabled the development of AI emulators for atmospheric and ocean circulation prediction \cite{weyn2021sub, clark2025ace2, dheeshjith2025samudra}. In particular, approaches based on time-series forecasting, large language models, and operator learning have been adapted to high-volume two and three-dimensional meteorological and climate datasets \cite{weyn2021sub, dheeshjith2025samudra}. These AI emulators also leverage GPU-accelerated matrix computations to support deterministic and ensemble forecasting on both edge and high-performance computing platforms. Many models are trained on publicly available global and regional datasets, including ERA5, IFS, and HRRR, allowing them to capture complex land–sea–atmosphere interactions and support applications such as mesoscale convective system and tropical cyclone prediction. In contrast, another class of AI emulators \cite{chapman2025camulator,crossouard2026contribution} is trained using idealized general circulation configurations, such as aquaplanet \cite{medeiros2016reference} and the Atmospheric Modeling Intercomparison Project (AMIP) \cite{gates1999overview}, providing an alternative to training on global and regional observational or reanalysis datasets.\\

Many AI emulators have achieved forecast skills comparable to the state-of-the-art physics-based models for medium-range to sub-seasonal forecast horizons \cite{weyn2021sub, pathak2022fourcastnet, crossouard2026contribution}. Meteorological agencies like ECMWF and NOAA have created frameworks like anemoi and NOAA-Eagle to develop in-house capabilities to generate and deploy AI emulators for weather forecasting applications. Therefore, the accuracy of AI emulators to predict general circulation and extreme weather events is crucial to support their transition to operation and promotion as a decision-making support system. However, training and testing of emulators have a large computational load that serves as a bottleneck to scaling AI emulators for (i) high-spatial resolution modeling, and (ii) out-of-the-box solutions on resource-constrained environments. Therefore, improvements in model memory footprint, inference time, and power consumption improve the feasibility of scaling the emulators. \\

To achieve accelerated inference with lower memory footprint and energy usage, a principled methodology is to reduce the precision of the operands in a machine learning model to a representation requiring less number of bits. This is known as quantization, where 32-bit floating-point (FP32) or 16-bit Brain Floating Point (BF16) parameters are mapped to lower-bit representations, reducing model memory and energy requirements~\cite{nagel2021white}. For example, FP4 provides approximately $8\times$ lower memory usage than FP32 while retaining floating-point dynamic range~\cite{liu2023llm,rouhani2023microscaling}. Well-designed FP4 quantization schemes can achieve accuracy comparable to BF16~\cite{wangoptimizing}. While quantization can reduce the memory footprint and energy requirements, it also increases the throughput since the arithmetic operations for matrix multiplications can be performed with a lesser amount of operations per bit. As a result, recent state-of-the-art GPUs such as the NVIDIA Blackwell and Rubin natively support computations with FP4 weights. \\

Quantization of DL models can be categorized into two forms, namely Post-Training Quantization (PTQ), and Quantization Aware Training (QAT). PTQ performs the mapping of full-precision model parameters to lower precision after the model has been trained in full-precision. QAT quantizes and dequantizes the model parameters while during the optimization steps in training to address the effect of quantization errors. At the system-deployment level, PTQ is preferred since it is fast, cheap, and requires no access to the original heavy training pipeline or large datasets. While QAT can yield higher accuracy, PTQ allows a finished, pre-trained model to be instantly compressed using only a tiny calibration set. \\

The representation of bits of input values, weights, and activations within a DL architecture governs the precision and accuracy of all output values. However, the quantization of different components can benefit DL models by reducing memory footprint and accelerating matrix multiplications, leading to high-throughput computations. Reducing precision in physics-based models has improved computational efficiency with insignificant degradation in forecast skill score \cite{chantry2019scale}. In the case of GCMs like OpenIFS and non-linear systems like the Lorenz 1963 and 1996 models \cite{jeffress2017bitwise, vavna2017single}, single floating-point precision contains sufficient information to overcome the uncertainties with precision loss and unresolved subgrid-scale processes. Quantization has been widely studied in computer vision and LLMs, focused on hardware acceleration and edge computing applications. Most PTQ and QAT training applications have focused on single-step inference. In this study, we implement PTQ algorithms to two different DL architectures to investigate the model optimization for weather forecasting applications.  \\

The contributions of this work are as follows:

\begin{itemize}
\item Post-training quantization (PTQ) is implemented in pre-trained AI emulators to optimize matrix multiplication for deterministic forecasting, spanning multiple weight and activation bit quantization across two different architectures: a convolutional U-Net (DLWP) and an AFNO-based vision transformer (FCN) 
\item We characterize quantization error under autoregressive
inference. Prior PTQ studies in vision and language evaluate
single-step prediction; we show how quantization error compounds across recursive steps in a non-linear system.
\end{itemize}

In Section 2, we describe the methodology to implement PTQ using NVIDIA Earth2Studio, forecast skill evaluation, and a list of bit representations and quantization algorithms. In Section 3, we discuss the results of inference runs including spatial maps and NWP skill scores. Section 4 provides a summary of the study followed by appendices describing the implementation.

\section{Methodology}\label{sec3}

\subsection{PTQ Workflow}

\begin{figure*}[ht]
\centering
\includegraphics[width=\linewidth]{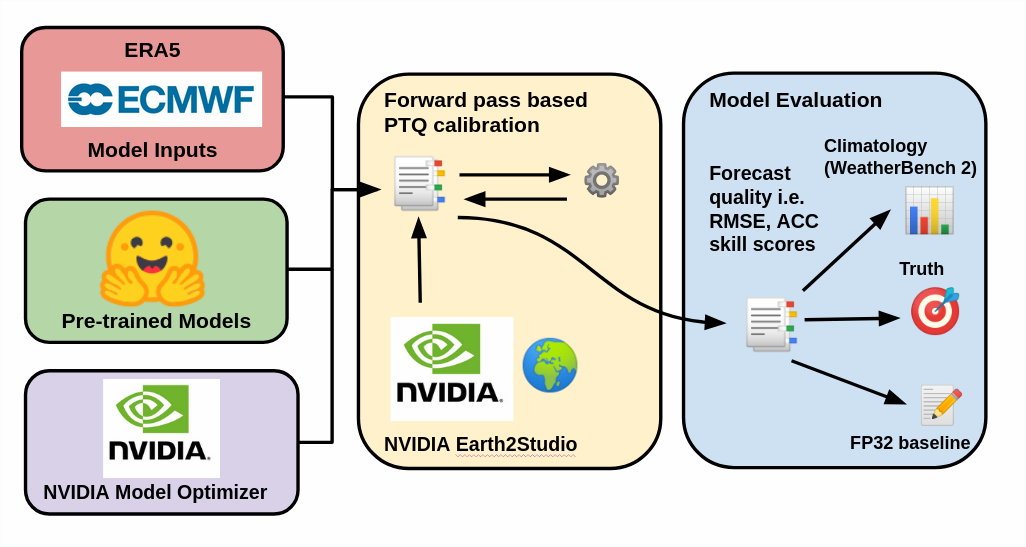}
\caption{Illustration of the PTQ implementation and model evaluation in DLWP and FCN models using NVIDIA Earth2Studio and Model Optimizer (ModelOpt). Earth2Studio calls pre-trained checkpoints for each model available on Hugging Face. The full precision models undergo weight and activation quantization based on inference based calibration. ECMWF ERA5 serve as the initial conditions to drive emulator inferences and evaluate their performance as a function of forecast hour. All PTQ experiments are executed using the configuration settings provided by ModelOpt.}
\label{fig:fig1}
\end{figure*}

Figure \ref{fig:fig1} illustrates the workflow for implementation of PTQ to pre-trained AI emulators, and the overall setup for model evaluation. We use the NVIDIA Earth2Studio toolkit to conduct forecast experiments and calibrate pre-trained AI emulators. Earth2Studio accesses the model checkpoints available on HuggingFace, and provides an interface to use global scale atmospheric dynamics data from ECMWF ERA5 as input to the AI emulators.  PTQ is implemented using the NVIDIA Model Optimizer package \cite{nvidia-modelopt} that leverages the PyTorch-based backend for DL emulators. ModelOpt can be configured to execute custom quantization strategies based on the treatment of weights and activations in different layers of the emulator architectures. We test the effect of both weight and activation quantization on key layers. The PTQ configurations are evaluated against the reference truth, full precision baseline, and climatology. Details of the model architectures, PTQ configuration, and model evaluation setup are presented in following subsections. 

\subsection{Model and PTQ Configurations}

We choose two DL emulators based on the model architectures: Deep Learning Weather Prediction (DLWP) \cite{weyn2021sub} and FourCastNet (FCN) \cite{pathak2022fourcastnet}. Characterizing the emulator performance subject to different PTQ precision and algorithms provides a robust understanding of forecast quality and predictability horizon. \\

DLWP uses a Convolutional Neural Network in a U-Net architecture on a cubed-sphere grid \cite{weyn2021sub}. The skip connections enabled by U-Net ensure preservation of high-resolution information across the layers \cite{ronneberger2015u}. It recursively predicts the atmospheric variables at a six-hour time resolution: geopotential at 4 different pressure levels 300 hPa (z300), 500 hPa (z500), 700 hPa (z700) and 1000 hPa (z1000); 2 m temperature (t2m); temperature at 850 hPa (t850); and total column water vapor (tcwv). DLWP is trained on ERA5 reanalysis \cite{hersbach2020era5} dataset remapped to 1.4$^{o}$ effective resolution at the equator.\\

On the other hand, FCN is based on a vision transformer (ViT) architecture with a Fourier transform-based token-mixing scheme. This architecture enables processing of high-resolution spatio-temporal data required for NWP applications. We specifically look at the geopotential variables, 2 m temperature, 10 m wind speeds, and total column water vapor to track the changes in forecast skill scores. Table 1 provides a short summary of the DL emulators, the model architectures, input dataset, and output variables considered for this study. The units used in this study are kept consistent for the variables.\\

\begin{table*}[t]
\centering
\caption{A summary of the deep learning atmospheric dynamics emulators, the back-end architecture, input data type and output variables considered for evaluation in this study.}
\label{tab:models}
\begin{tabularx}{\textwidth}{l l l X}
\toprule
Model Name & Architecture & Input Type & Output Variables \\
\midrule
DLWP & U-Net (cubed sphere) & ECMWF ERA5 & $z_{300}$, $z_{500}$, $z_{700}$, $z_{1000}$, $t_{850}$, $t_{2m}$, $tcwv$ \\
FCN  & AFNO (ViT backbone)  & ECMWF ERA5 & $z_{500}$, $z_{1000}$, $t_{850}$, $t_{2m}$, $u_{10m}$, $v_{10m}$, $tcwv$ \\
\bottomrule
\end{tabularx}
\label{tab1}
\end{table*}

A wide range of precision levels are tested, considering floating-point (FP) FP32 \cite{8766229} as the baseline, namely: INT8, INT4 and INT2 \cite{nagel2021white, jacob2018quantization}. For the A32 configurations, activations are kept at full precision to preserve the statistics of the pre-trained checkpoints. Two special types of quantization are implemented for lower integer precision, i.e., INT8 SmoothQuant \cite{xiao2023smoothquant} and INT4 Activation-aware Weight Quantization (AWQ) \cite{lin2024awq}. The weight and activation conversions are based on calibration of each model based on a finite number of forward passes. Following standard practice in PTQ, the input layer, output layer, and normalization are kept at full precision. The input layer is subjected to a large dynamic range of physical variables while the output layer directly influences forecast accuracy. The set of PTQ experiments with their respective descriptions for weight and activation bit representation is summarized in Table 2. \\

\begin{table*}[t]
\caption{A list of the PTQ methods described with respective weight precision, activation precision and notes for the quantization algorithm applied to pre-trained checkpoints on full-precision based DLWP and FCN emulators}
\label{t1}
\centering
\footnotesize
\setlength{\tabcolsep}{4pt}      
\renewcommand{\arraystretch}{1} 
\begin{tabular*}{\columnwidth}{@{\extracolsep{\fill}}lccc@{}}
\toprule
Name & Weight & Activation & Notes \\
\midrule
FP32 & FP32 & FP32 & Full precision baseline \\
W8A8 & INT8 & INT8 & Tensor weight scaling-based quantization \\
W8A32 & INT8 & FP32 & INT8 weights with FP32 activations \\
W4A32 & INT4 & FP32 & INT4 weights with FP32 activations \\
W2A32 & INT2 & FP32 & INT2 weights with FP32 activations \\
INT8 SmoothQuant & INT8 & INT8 & Activation smoothing using SmoothQuant \\
INT4 AWQ & INT4 & FP32 & Activation-aware Weight Quantization \\
\bottomrule
\end{tabular*}
\end{table*}

\subsection{Evaluation setup}

We evaluate the reduced emulators for short-range forecasting on four distinct dates corresponding to various seasons of the year 2020 (DLWP) and 2022 (FCN), i.e., the 1st date of January, April, July, and October. The 1st January initial condition is used to calibrate the PTQ over an inference window of 48 hours comprising 8 inference steps. For evaluation purposes, the inference window is kept at 10 inference steps equal to a 60-hour lead time.  The collection of inferences provides an estimation of intra-annual uncertainties in the PTQ approach. For a collection of output variables in DLWP and FCN, the forecast quality is evaluated using two traditional skill metrics i.e. root mean square error (RMSE) and anomaly correlation coefficient (ACC) \cite{palmer1988prediction}. \\

\begin{equation}
\label{eq1}
\mathrm{RMSE} = \sqrt{\frac{1}{N}\sum_{i=1}^{N} w_{i}(x_{f,i} - {x}_{t,i})^2}
\end{equation}

\begin{equation}
\label{eq2}
w_{i} = \frac{cos(\theta_{i})}{\frac{\sum_{i=1}^{N_{lat}}cos(\theta_{i})}{N_{lat}}}
\end{equation}

\begin{equation}
\label{eq3}
\mathrm{ACC} = \frac{\sum_{i=1}^{N} (x_{f,i} - x_{c,i})(x_{t,i} - x_{c,i})}{\sqrt{\sum_{i=1}^{N}(x_{f,i} - x_{c,i})^{2} \sum_{i=1}^{N}(x_{t,i} - x_{c,i})^{2}}}
\end{equation}

Equations \ref{eq1} - \ref{eq3} show the mathematical description of RMSE and ACC skill scores. Here, $x_{f}$ represents the forecast value, $x_{c}$ is the climatology value and $x_{t}$ is the reference truth based on atmospheric reanalysis, forecast analysis or full precision inference. The index $i$ represents the index of a grid point as the skill scores are summed up over the domain with N grid points equal to product of latitude and longitude bins. RMSE is weighted by latitude \\

The RMSE and ACC skill scores collectively provide a holistic perspective on the forecast accuracy and horizon of predictability. RMSE scores are compared to the reference forecast truth as well as the FP 32 baseline. DLWP and FCN models are initialized using ERA5 inputs. We use the WeatherBench 2 \cite{rasp2024weatherbench} ERA5 Climatology as the reference for computing the ACC skill scores. Each set of PTQ inferences are tested on two different GPU architectures: NVIDIA Volta (V100) \cite{martineau2018benchmarking} and Ampere (A100). We run the simulations on the National Center for Atmospheric Research (NCAR) Casper supercomputer using a Python-based virtual environment and PBS job submission pipeline.

\section{Results}\label{sec4}

\subsection{Skill Scores and Forecast Quality}

\begin{figure*}[ht]
\centering
\includegraphics[width=\linewidth]{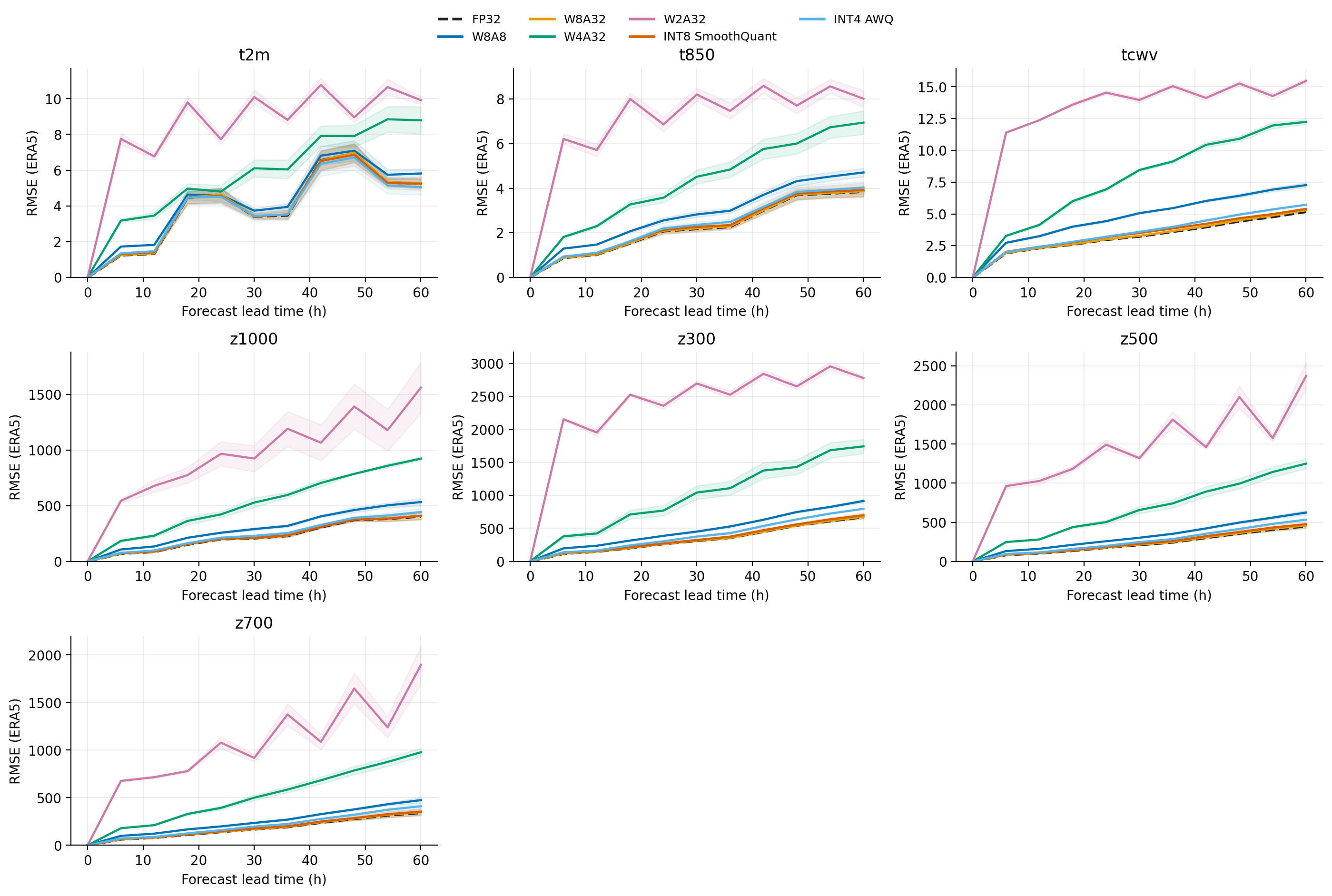}
\caption{RMSE forecast error as a function of lead time for DLWP inference runs relative to ERA5. The growth of error in PTQ configurations are compared against the full precision (black dotted) baseline in the above panels: z500 ($\mathrm{m^2\,s^{-2}}$), z300 ($\mathrm{m^2\,s^{-2}}$), z700 ($\mathrm{m^2\,s^{-2}}$), z1000 ($\mathrm{m^2\,s^{-2}}$), t850 (K), t2m (K) and tcwv ($\mathrm{kg\,m^{-2}}$). RMSE errors also highlight the variations within a lead time for selected initial condition representing seasonal variability in forecast error.}
\label{fig:fig2}
\end{figure*}

We present the RMSE and ACC skill scores of emulator inferences as representative results of the PTQ experiments. Figure \ref{fig:fig2} illustrates the evolution of latitude-weighted RMSE metrics relative to ECMWF ERA5 reanalysis. Differences from ERA5 represent deviations from the baseline truth based on state-of-the-art atmospheric reanalysis.  The FP32 RMSE relative to ERA5 show a slow linear increment along the set of inference runs, except for t850 and t2m. The stepwise increment in RMSE likely represents the diurnal cycle in atmospheric temperature values. A similar trend is also observed for most PTQ experiments. \\

\begin{figure*}[ht]
\centering
\includegraphics[width=\linewidth]{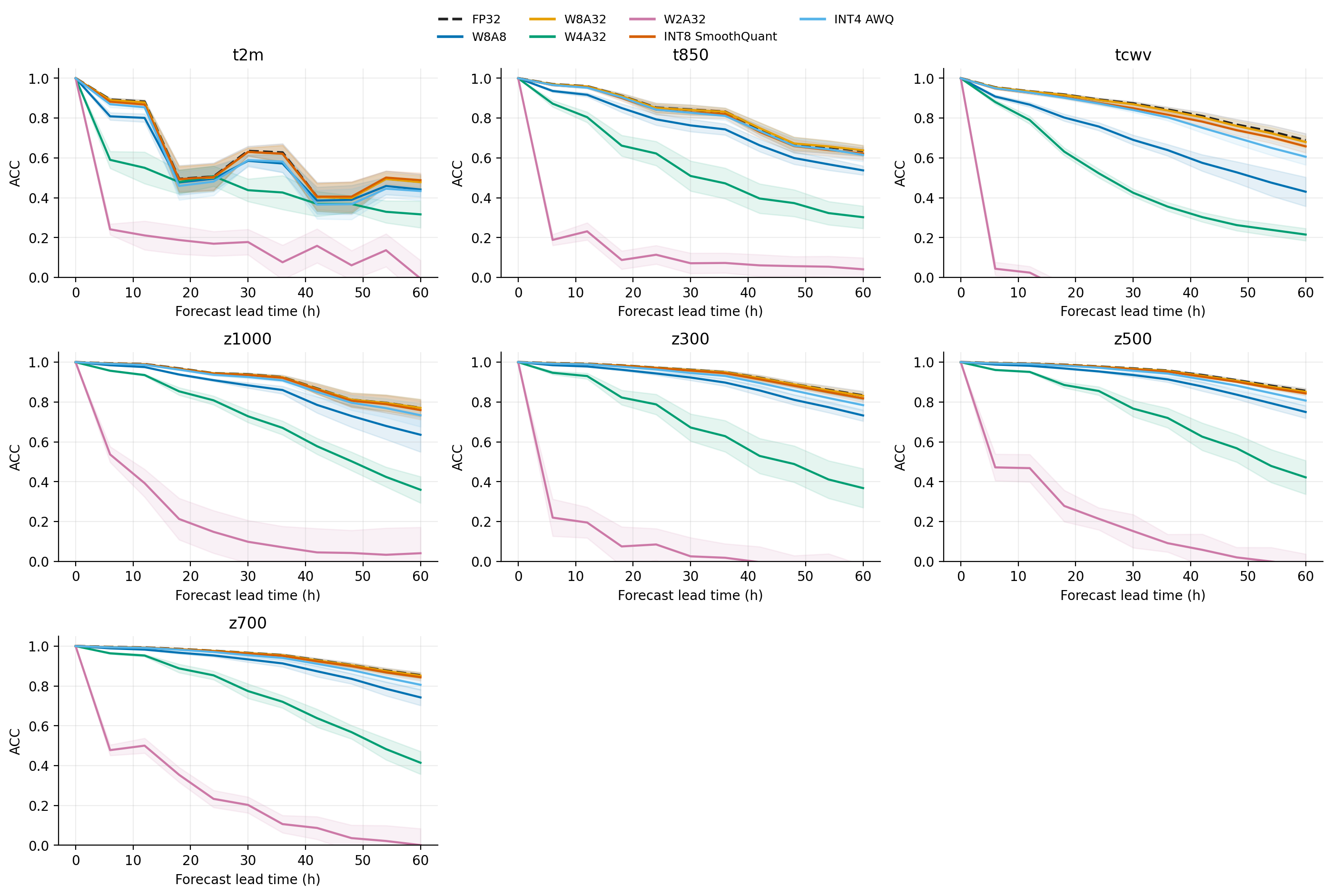}
\caption{ACC forecast skill score as a function of forecast lead time in DLWP inference runs relative to ERA5 as truth and WeatherBench2 climatology. Change in ACC skill highlights the quality of PTQ experiments relative to the FP32 baseline.}
\label{fig:fig3}
\end{figure*}

The ACC skill score provides additional insights on forecast relative to ERA5 reanalysis and ERA5 climatology. Figure \ref{fig:fig3} presents the time evolution of ACC over the forecast horizon for geopotential and atmospheric temperature variables. The 500 hPa geopotential ACC is used by ECMWF and NCEP for validating their forecast quality relative to the truth and climatology. For geopotential variables, the ACC scores drop after 36 hours and reach a value of 0.55 to 0.8 for the best-performing cases. A value of 0.6 or greater in geopotential ACC represents a good quality of forecast. Temperature and tcwv ACC skill scores for best PTQ experiments closely follow the trends with FP32 full precision inferences.\\

\begin{figure*}[ht]
\centering
\includegraphics[width=\linewidth]{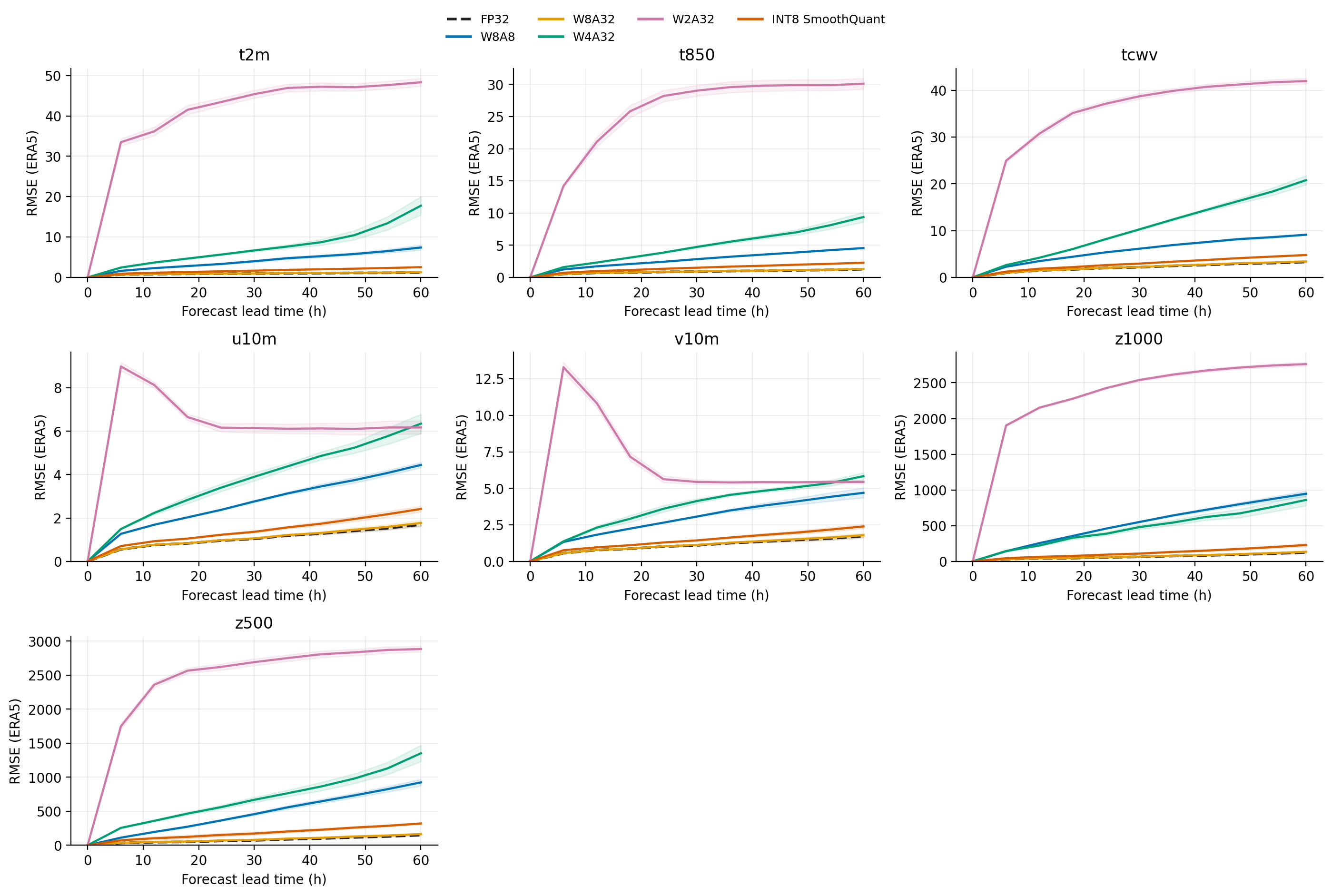}
\caption{RMSE forecast error as a function of lead time for FCN inference runs relative to ERA5. The growth of error in PTQ configurations are compared against the full precision (black dotted) baseline in the above panels: z500 ($\mathrm{m^2\,s^{-2}}$), z1000 ($\mathrm{m^2\,s^{-2}}$), t850 (K), t2m (K), tcwv ($\mathrm{kg\,m^{-2}}$), u10m ($\mathrm{m\,s^{-1}}$) and v10m ($\mathrm{m\,s^{-1}}$). RMSE errors also highlight the variations within a lead time for selected initial condition representing seasonal variability in forecast error.}
\label{fig:fig4}
\end{figure*}

\begin{figure*}[ht]
\centering
\includegraphics[width=\linewidth]{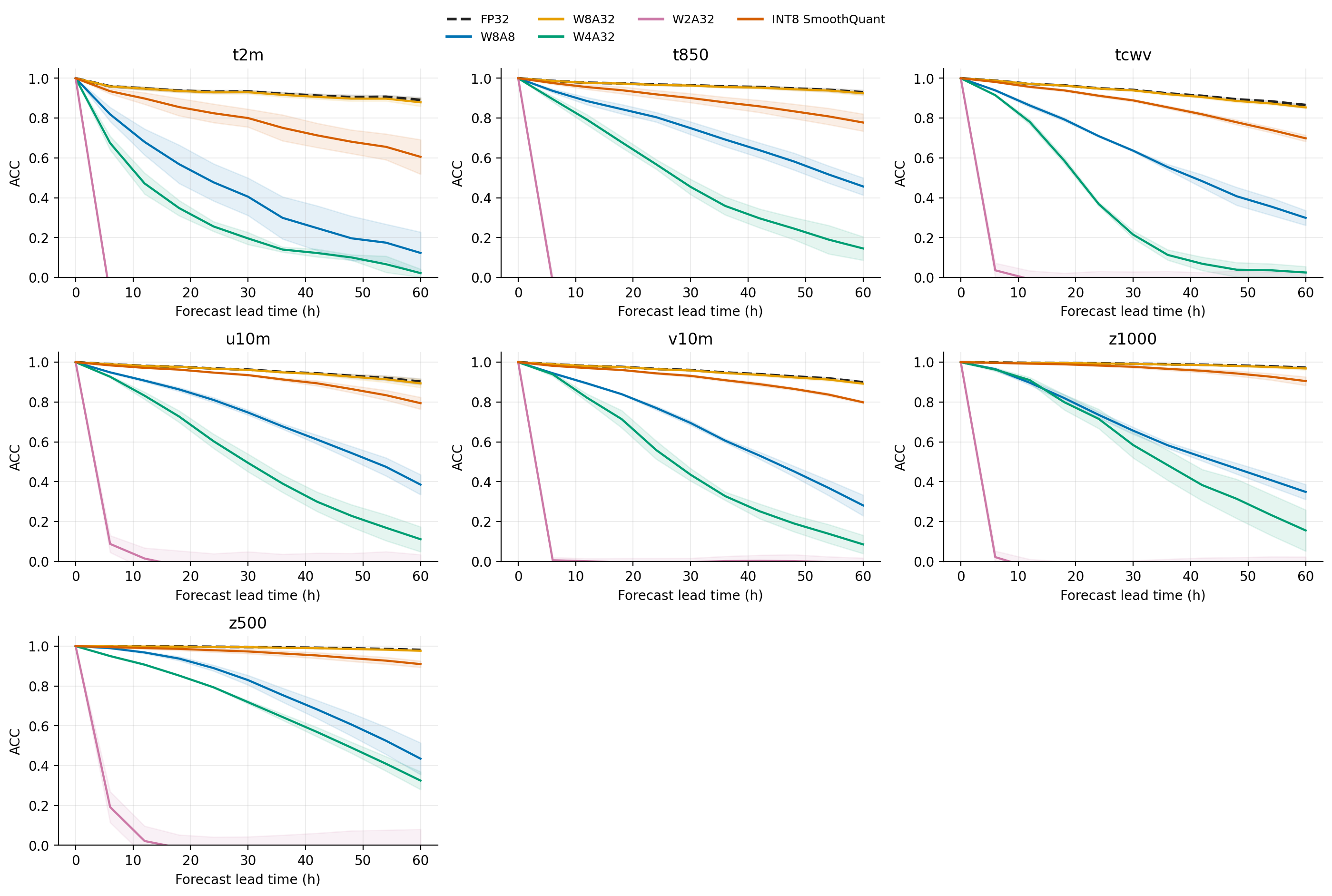}
\caption{ACC forecast skill score as a function of forecast lead time in FCN inference runs relative to ERA5 as truth and WeatherBench2 climatology. Change in ACC skill highlights the quality of PTQ experiments relative to the FP32 baseline.}
\label{fig:fig5}
\end{figure*}

PTQ experiments show a similar trend in FCN inferences driven by ERA5. Figure \ref{fig:fig4} shows that the FCN RMSE scores for the geopotential, atmospheric temperature, and wind speed variables show a relatively larger variation between the four initial conditions post the 24-hour window. The W8A32 and SmoothQuant configurations provide the overall best performance over the 60-hour window. As the RMSE skill scores are calculated relative to ERA5, the performance highlights the overall variation between different seasons. Similar variations are also exhibited by ACC skill score relative to WeatherBench2 Climatology and ERA5. Figure \ref{fig:fig5} shows that the z500 ACC also show a significant variation within the 60 hour lead time. However, the overall ACC values lie above 0.65 highlighting the capability of FCN to produce high-quality short-term forecast. The collective RMSE and ACC scores and their evolution with forecast time demonstrates the effectiveness of PTQ algorithms in DL emulators. These experiments show that changes in weight and activation precision provides qualitatively meaningful forecast over a short-time forecast horizon. \\

\subsection{Spatial Maps of Forecast Variables}

\begin{figure*}[ht]
\centering
\includegraphics[width=1.1\linewidth]{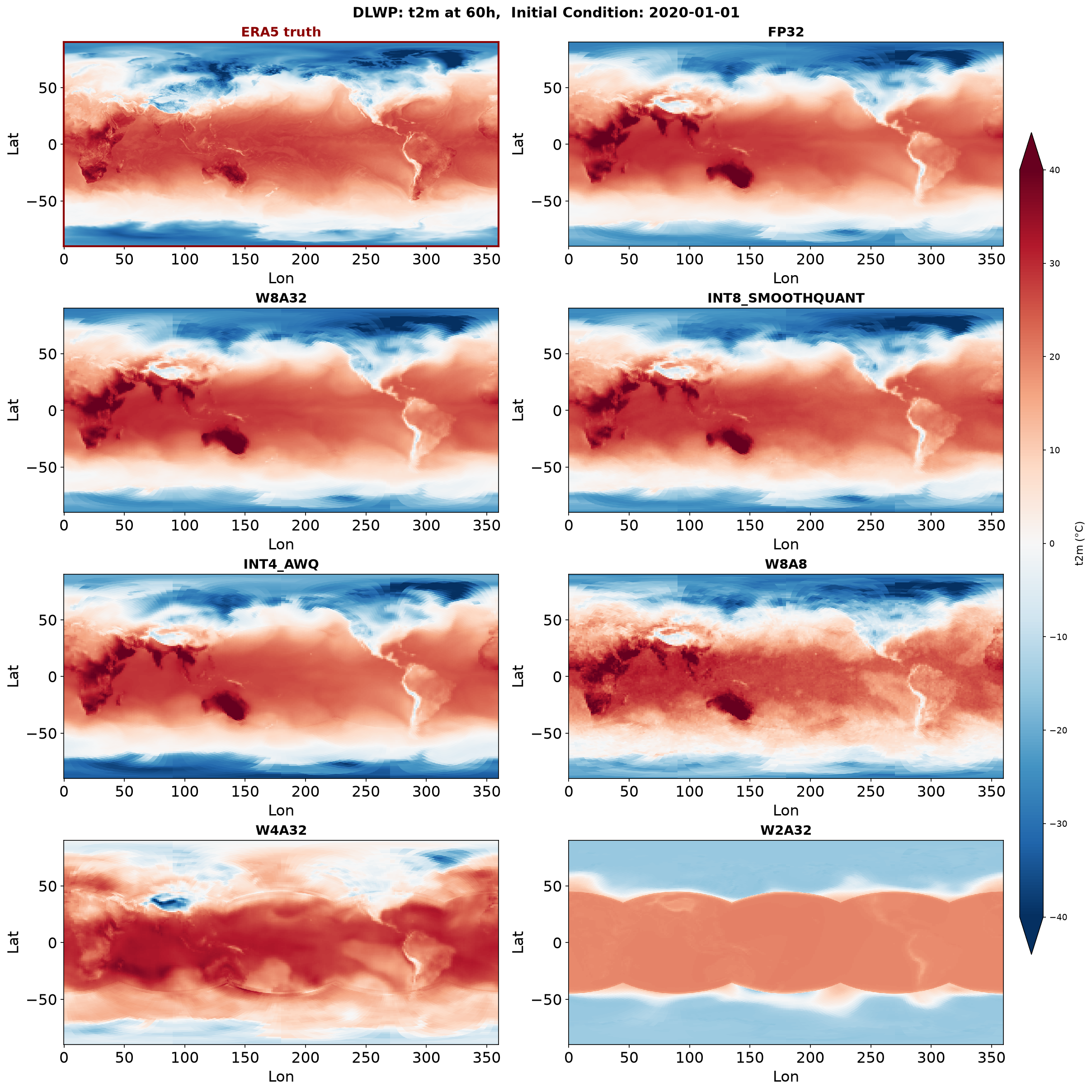}
\caption{Two-dimensional spatial maps for 2 m Temperature  at the end of 60-hour inference runs for the DLWP model calibrated to different PTQ configurations. ERA5 is the reference truth, and FP 32 is the reference for precision for DLWP capability.}
\label{fig:fig6}
\end{figure*}

\begin{figure*}[ht]
\centering
\includegraphics[width=1.1\linewidth]{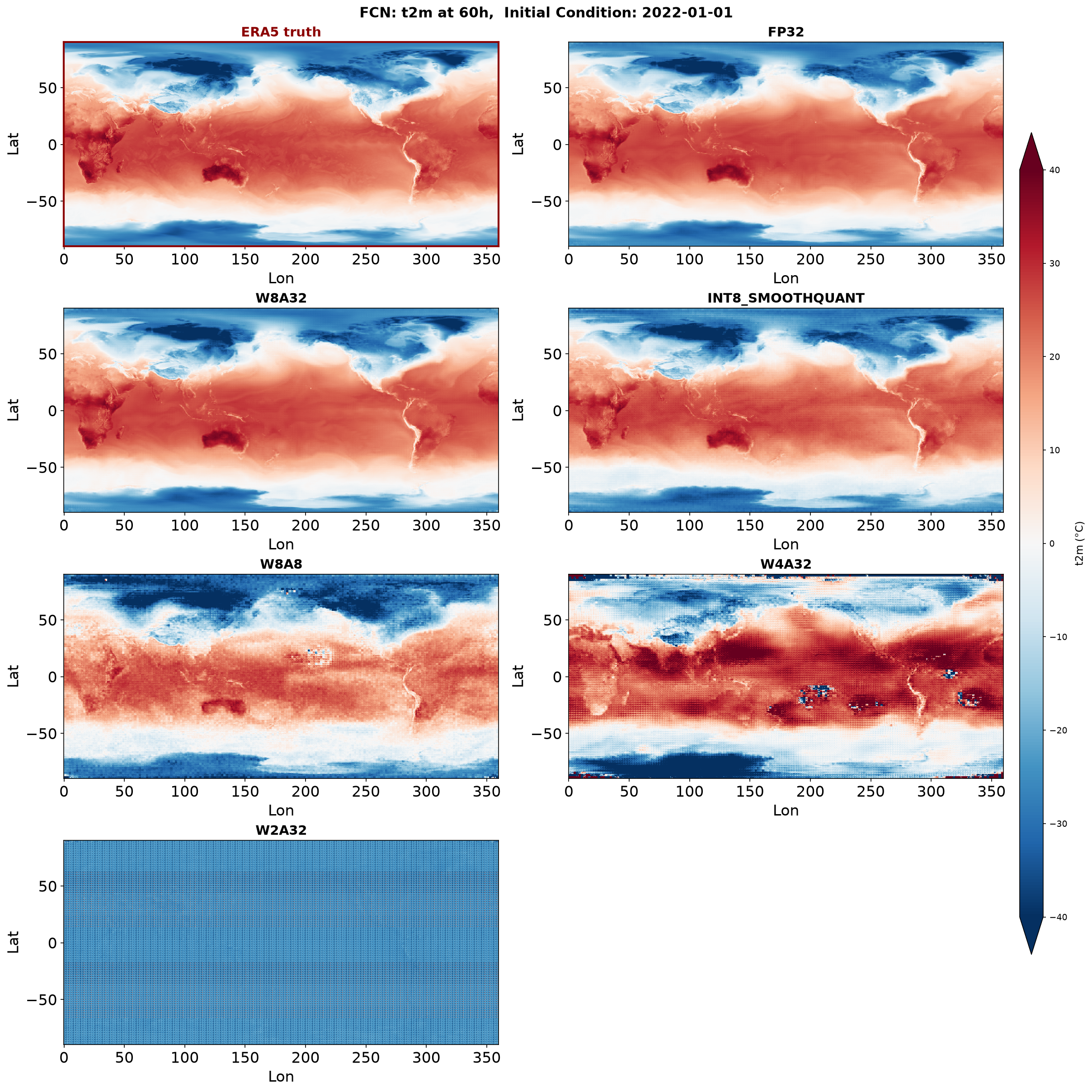}
\caption{Two-dimensional spatial maps for 2 m Temperature at the end of 60-hour inference runs for the FCN model calibrated to different PTQ configurations. ERA5 is the reference truth, and FP32 is the reference for precision for FCN capability.}
\label{fig:fig7}
\end{figure*}

Here, we present the spatial maps of 2 m air temperature as forecast variables at the end of 60-hour inference runs. The maps are compared against the reference truth and FP 32 full precision inference. Spatial variations in air temperature and degradation relative to the truth and FP32 highlights the precision loss, and biases in forecast output. Figures \ref{fig:fig6} and \ref{fig:fig7} summarize the qualitative effect of PTQ on DLWP forecast. Reducing the weight and activation precision introduces coarse features in the forecast along with biases inherited from the model itself. Special PTQ algorithms like AWQ and SmoothQuant preserve the spatial quality relative to FP32 inference. A special case of PTQ based on INT2 quantization in weights highlights features consistent with the underlying cubedsphere grid in DLWP. \\

Specific hot and cold biases are observed in the t2m field depending on the land-sea boundary and geographical locations. A detailed comparison with ERA5 as a baseline truth reveals the magnitude and spatial coverage of the hot and cold bias. The quantized inferences capture the spatial trends relative to the FP 32 inference. Reducing the precision inevitably leads to less accurate forecast values. However, intermediate precision values with special quantization algorithms preserve the spatial features and forecast skills within the 60-hour lead time. \\

The FCN results are similar to the DLWP PTQ inferences. Spatial maps show conservation of forecast skill against precision loss for multiple configurations like W8A32 and INT8 SmoothQuant. The INT2 (W2A32) case shows periodic patterns in spatial maps reflecting the Fourier transform-based mixing of embeddings. Therefore, low precision inferences are found to retain information from underlying coordinate system or embeddings. These spatial maps demonstrates the capability of DL emulators to produce good quality forecast despite inherent biases in output fields, and precision loss over forecast steps.

\subsection{Hardware Metrics}

\begin{figure*}[ht]
\centering
\includegraphics[width=\linewidth]{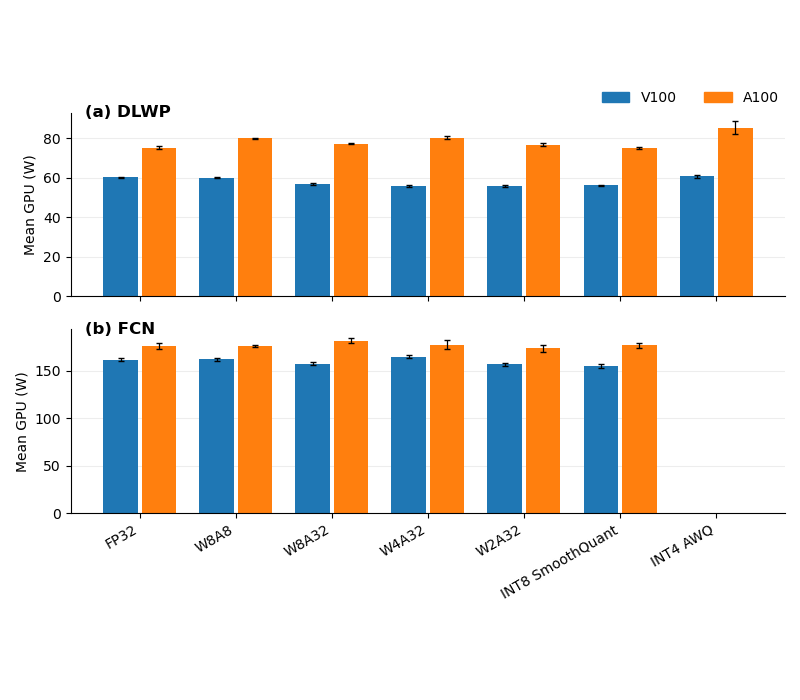}
\caption{Mean power consumption (W) of model inferences on V100 and A100 GPU nodes for different PTQ configurations implemented in (a) DLWP and (b) FCN models through the NVIDIA Earth2Studio package}
\label{fig:fig8}
\end{figure*}

Quantization in AI models is expected to make them efficient for low-resource computing platforms. Memory footprint, power consumption, and GPU inference times are some important parameters that quantify the efficiency of the models undergoing PTQ. For our experiments, the memory footprint does not change as ModelOpt applies a fake quantization to the pre-trained models. The hardware-related metrics are recorded for a set of 5 inference runs for each experiment on a given GPU node. We query the GPU power consumption during the inference runs using the PyNVML tool based on the NVIDIA Management Library (NVML). Figure \ref{fig:fig8} summarizes the mean power consumption and its variation over a forecast inference for DLWP and FCN models. It is not found to change significantly between different PTQ configurations. However, the mean power consumption during a forecast inference is dependent on the model and GPU architecture. Systematically, the V100 consumes less power relative to the A100 nodes over all configurations. \\

\begin{figure*}[ht]
\centering
\includegraphics[width=\linewidth]{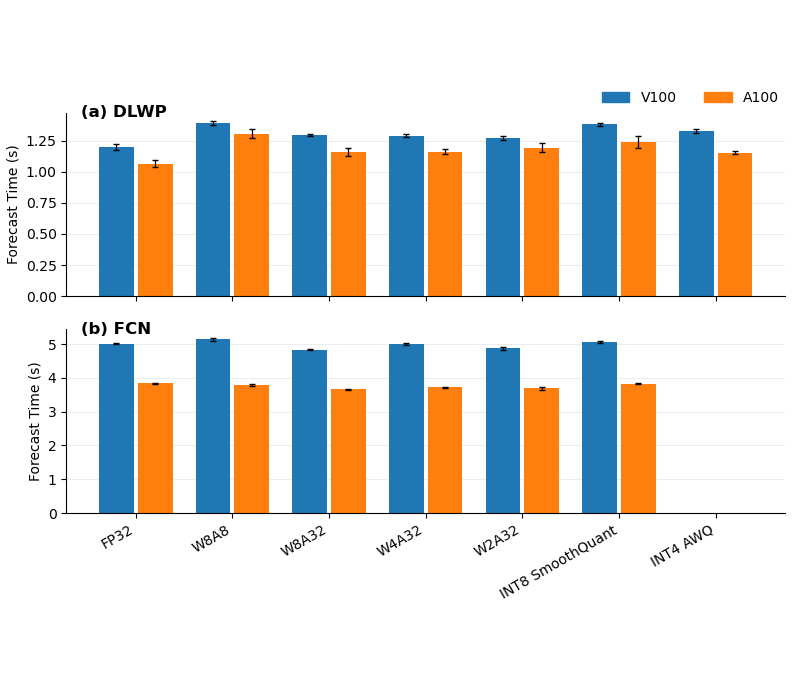}
\caption{Total inference time per forecast (s) on V100 and A100 GPU nodes for different PTQ configurations implemented in (a) DLWP and (b) FCN models through NVIDIA Earth2Studio package}
\label{fig:fig9}
\end{figure*}

Figure \ref{fig:fig9} shows the total inference time per forecast for the two DL emulators. The inferences systematically perform better on A100 relative to V100 in accordance with hardware performance expectations. PTQ experiments in both models introduce additional forecast time relative to FP32 baseline. This could be attributed to interconversion between each layer and fake quantization to achieve full-precision output at each step. FCN shows more consistent performance during PTQ relative to DLWP, however, DLWP computations are 2-4 times faster. Overall, the inference times are found to be hardware-dependent, and PTQ leads to minor increments in most of the cases.

\section{Summary and Conclusion}\label{sec6}

In this work, we implement PTQ-based model optimization of deep-learning-based global atmospheric dynamics models for weather prediction applications. NVIDIA Earth2Studio and ModelOpt provide a seamless way of implementing and testing different experiments focused on weight and activation quantization. The model evaluation setup demonstrates that applying quantization to pre-trained models can preserve model statistics at mixed precision involving INT8 and INT4 bit representation with full-precision activation. Evaluation of RMSE and ACC skill scores in DLWP and FCN provides a proof of concept for effective optimization of deep learning models for weather prediction.  As the inference runs are executed autoregressively, multi-step inferences are expected to degrade due to low bit representation. However, PTQ experiments are observed to provide qualitatively meaningful forecasts over short-range forecast horizons. \\

We observe that weight bit quantization plays a dominant role in forecast skill degradation relative to changes in activation bit quantization. Implementation of DL architecture-aware quantization like SmoothQuant and AWQ performs better than naive absolute-maximum-based PTQ calibration. The set of PTQ experiments serves as the first benchmark for an autoregressive DL emulator focused on weather forecasting applications to our knowledge. It can also be extended to implement QAT during training and fine-tuning of AI weather emulators. The physical interpretation and impact of quantization on DL emulators can be assessed using idealized test cases of general circulation. A hierarchical testing of DL emulators by training them on idealized general circulation cases could provide additional insights on the physical consistency of inferences produced by quantized emulators. \\

\section{Data and Code Availability}
NVIDIA Earth2Studio and ModelOpt are available as open source software accessible through GitHub \cite{Earth2Studio_Contributors_NVIDIA_Earth2Studio_2024, nvidia-modelopt}. The checkpoints for pre-trained models are available on HuggingFace, and they have been accessed through Earth2Studio. The code for executing the PTQ experiments, evaluating the model, and collecting hardware metrics are available in a separate GitHub repository \cite{ananyo_bhattacharya_2026_21969498}.

\section{Acknowledgment}

We would like to acknowledge computing support from the Casper system \url{https://ncar.pub/casper} provided by the NSF National Center for Atmospheric Research (NCAR), sponsored by the National Science Foundation (NSF).
\bibliography{agusample}
%
%


%
%
%
%
%

\appendix
\section{Appendix}
\subsection{Forecast Skill Relative to FP32}

\begin{figure*}[ht]
\centering
\includegraphics[width=\linewidth]{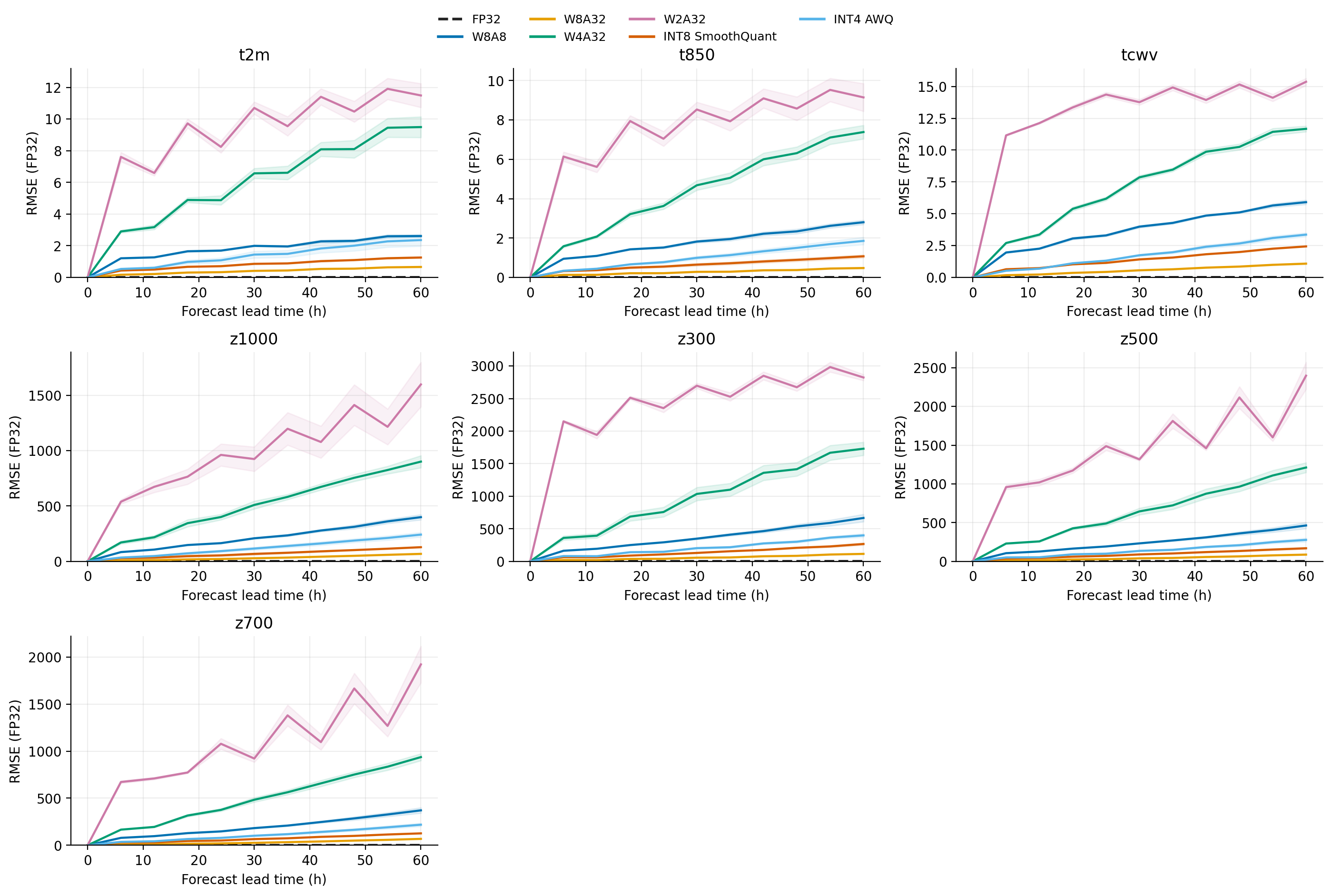}
\caption{RMSE error relative to FP32 (black dotted) inference as a function of forecast lead time for DLWP model configured to PTQ experiments. Growth in forecast error relative to FP32 highlights the degradation of forecast skill for different cases of weight and activation quantization.}
\label{fig:fig10}
\end{figure*}

The differences in output fields relative to the full precision baseline is an important factor in the characterization of PTQ experiments. Figure \ref{fig:fig10} traces the divergence of PTQ experiments in the DLWP model relative to the FP32 baseline. It represents the quality of PTQ experiments and the degradation in forecast quality relative to the best achievable model performance at full precision. It can be observed that across all variables, the SmoothQuant and AWQ-based calibrations perform better compared to absolute max-based calibration at lower bit representation for model weights. For geopotential variables, the RMSE is less than 500 m$^{2}$s$^{-2}$ for the best-performing PTQ configurations. Similarly, the RMSE in temperature values is found to be less than 2 K, highlighting the capabilities of quantization for global scale weather forecasting applications.\\

\begin{figure*}[ht]
\centering
\includegraphics[width=\linewidth]{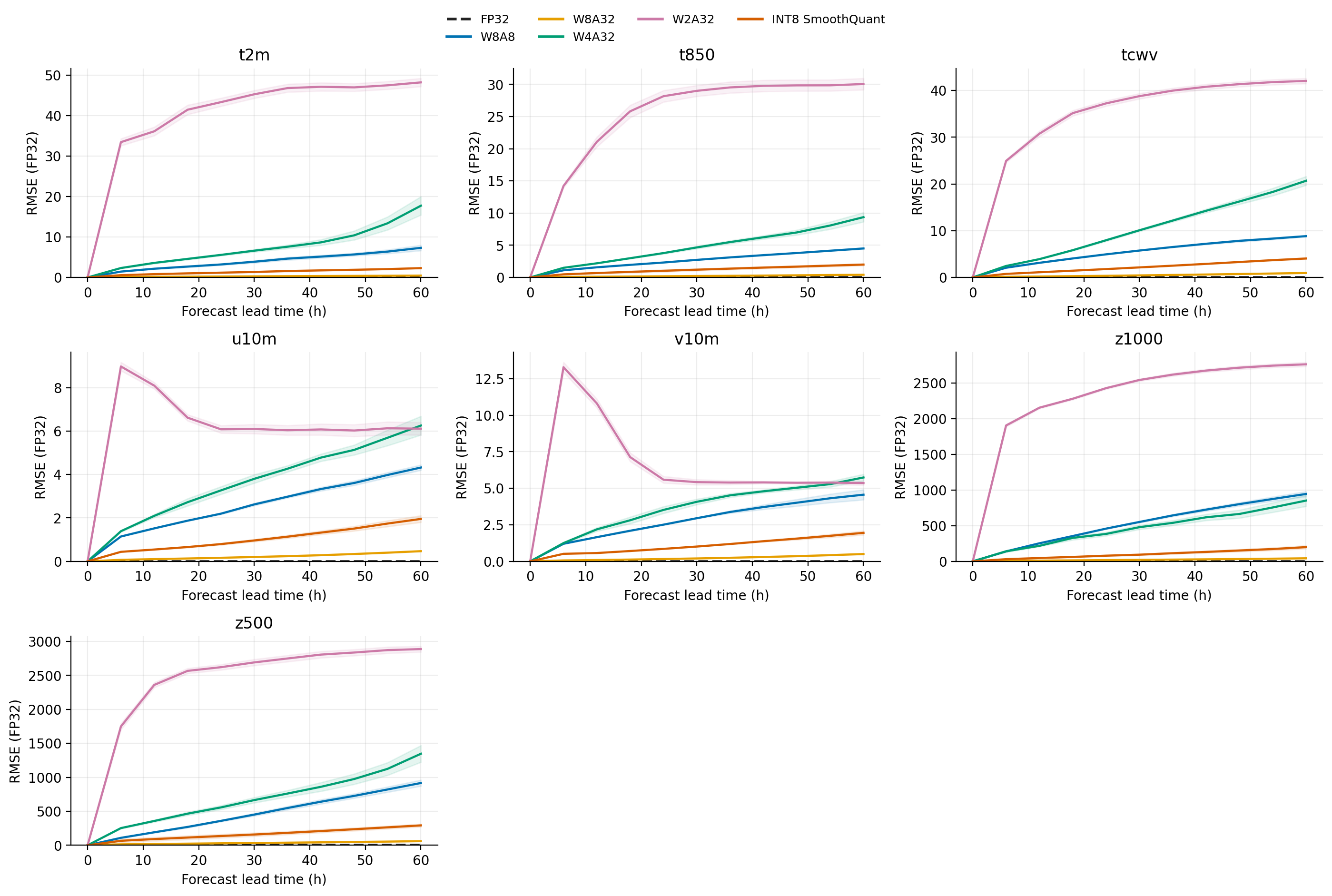}
\caption{RMSE error relative to FP32 (black dotted) inference as a function of forecast lead time for FCN model configured to PTQ experiments. Growth in forecast error relative to FP32 highlights the degradation of forecast skill for different cases of weight and activation quantization.}
\label{fig:fig11}
\end{figure*}

Figure \ref{fig:fig11} shows the time-dependent evolution of RMSE in FCN model inference runs. A significant amount of variation in RMSE skill scores is found in specific variables like z500, t850, t2m and u10m. On the other hand, z1000, v10m and tcwv do not show significant variation in RMSE for different initial conditions. The W2A32 skill scores show a non-monotonic evolution in wind speed variables, and overall do not provide a qualitatively meaningful forecast.

\subsection{Spatial Maps of Water Vapor}

\begin{figure*}[ht]
\centering
\includegraphics[width=1.1\linewidth]{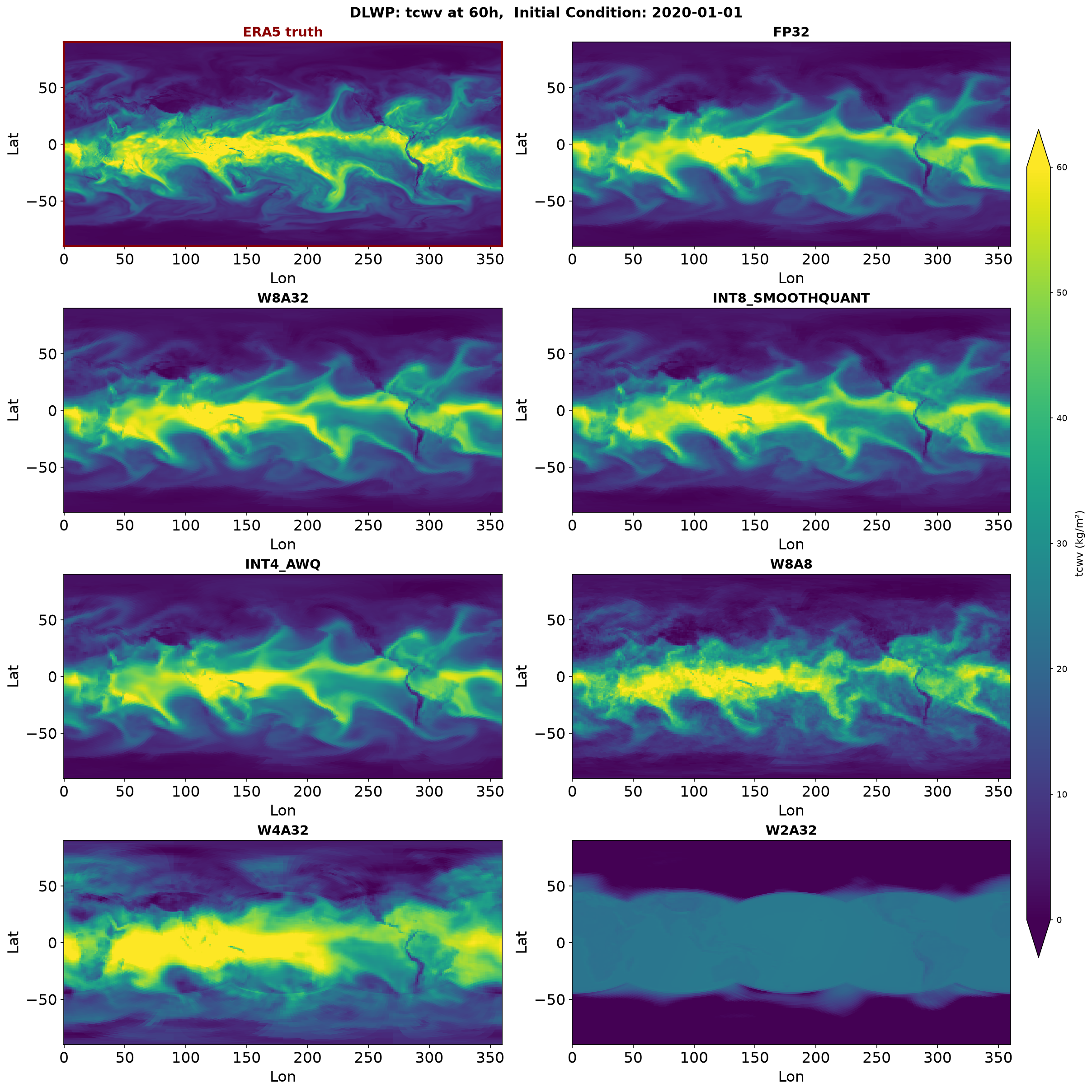}
\caption{Two-dimensional spatial maps for total column water vapor at the end of 60-hour inference runs for the DLWP model calibrated to different PTQ configurations. ERA5 is the reference truth, and FP 32 is the reference for precision for DLWP capability.}
\label{fig:fig12}
\end{figure*}

\begin{figure*}[ht]
\centering
\includegraphics[width=1.1\linewidth]{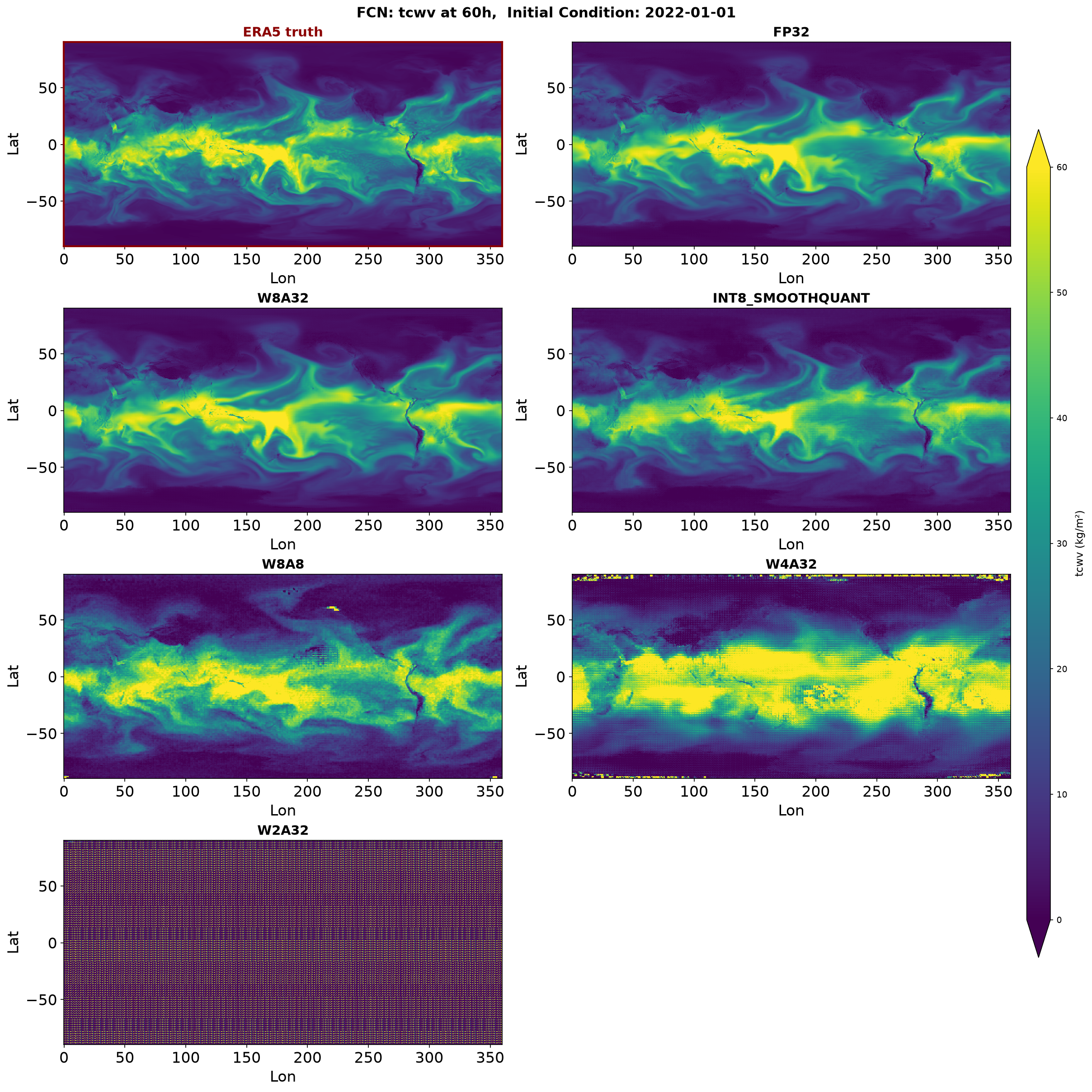}
\caption{Two-dimensional spatial maps for total column water vapor at the end of 60-hour inference runs for the FCN model calibrated to different PTQ configurations. ERA5 is the reference truth, and FP32 is the reference for precision for FCN capability.}
\label{fig:fig13}
\end{figure*}

The spatial maps of total column water vapor (tcwv) at the end of emulator inferences are presented in Figures \ref{fig:fig12}-\ref{fig:fig13}. Similar to the air temperature maps, the forecast quality degrades with a decrease in weight and activation bit representation. The output fields are smooth relative to the ERA5 truth values.

\end{document}